\documentclass{article}
\usepackage[utf8]{inputenc}
\usepackage{graphicx}
\usepackage{amsmath}
\usepackage{tabularx}
\usepackage{longtable}
\usepackage[top=3cm,bottom=5cm,left=4cm,right=3cm]{geometry}
\usepackage[toc,page]{appendix}
\usepackage{listings}
\usepackage{comment}
\usepackage{cancel}
\usepackage{float}
\usepackage{hyperref}
\usepackage{array}
\usepackage[center]{caption}

\title{ Geometric and Stiffness Modeling and Design of Calibration Experiments for the 7 dof Serial Manipulator KUKA iiwa 14 R820}
\author{Sami Sellami, Victor Massagué Respall }
\date{}

\begin{document}

\maketitle

\section{Introduction}
The present project deals with the elastostatic modeling and calibration experiment of spacial industrial manipulators using optimal selection of measurements pose \cite{klimchik2014geometrical}, for the calibration procedure, the optimal pose selection aims to the efficiency improvement of identification procedure for serial manipulators which reduces noise impact on the parameters identification precision, it is usually used for planar manipulators, our work is mainly to extend the approach for a more complicated manipulator in 3D space using a wise decomposition of the spacial manipulator into a set of serial sub-chains \cite{klimchik_math}, the optimal pose configuration is then used in the calibration procedure using the complete and irreducible model for the 7 dof serial manipulator \cite{klimchik2014geometric}.  The methodology is illustrated with the anthropomorphic industrial robot KUKA iiwa14 R820 for which, we performed the calibration and constructed the stiffness modeling using two different approaches namely VJM (Virtual Joint Modeling) and MSA (Matrix Structural Analysis).

\section{Related Works}
In contrast to previous works for calibration the proposed in \cite{klimchik2014geometrical} yields simple geometrical patterns that allow users to take into account the joint and workspace constraints and to find measurement configurations without tedious computations. The main theoretical results are expressed as a set of several properties and rules, which allow user to obtain optimal measurement configurations without any computation, just using superpositions and permutations of the proposed patterns. They presented an example for a 6 dof manipulator showing the efficiency. 

Similarly, \cite{wu2015geometric} proposes simple rules for the selection of manipulator configurations that allow the user to essentially improve calibration accuracy and reduce identification errors. The results are mainly for planar manipulators(two-, three- and four-link planar manipulators), but they can be used as a base for more complicated ones. The main contributions have been obtained for the planar case, the developed rule has been heuristically generalized for articulated robots. However, a strict theoretical proof of this approach remains unsolved for calibration of non-planar serial and parallel manipulators. 

In contrast to previous works, \cite{klimchik2014geometric} developed calibration technique based on the direct measurements only. To improve the identification accuracy, it is proposed to use several reference points for each manipulator configuration.The obtained theoretical results have been successfully applied to the geometric and elastostatic calibration of serial industrial robot employed in the machining work-cell for for aerospace industry.

\section{Methodology}
\subsection{Elastostic Modeling:}
\begin{figure}[H]
    \centering
    \includegraphics[width=\linewidth]{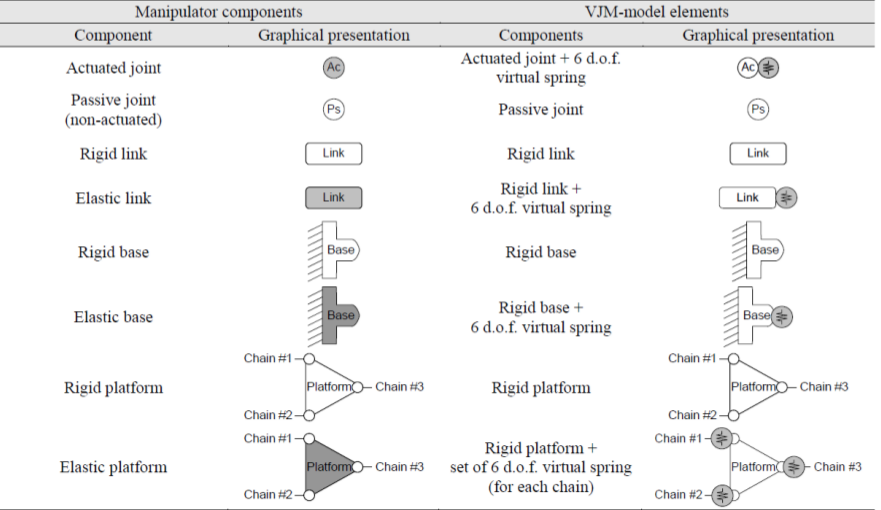}
    \caption{VJM base modeling of manipulator components}
    \label{fig:vjm_modeling}
\end{figure}

\paragraph{VJM model:}
To build the extended stiffness model of the KUKA iiwa robot, we are going to use the simplifications shown in Figure \ref{fig:vjm_modeling}

\begin{figure}[H]
    \centering
    \includegraphics[width=\linewidth]{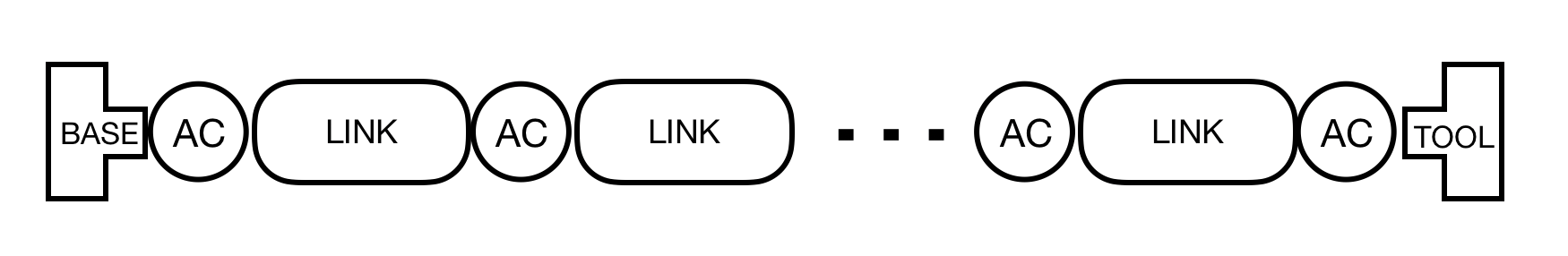}
    \caption{Robot model approximation of Kuka iiwa 14 R820}
    \label{fig:my_label}
\end{figure}

In the case of our manipulator, we have 6 elastic links and 7 actuated joints, the robot model approximation is shown in Figure \ref{fig:my_label}

\begin{figure}[H]
    \centering
    \includegraphics[width=\linewidth]{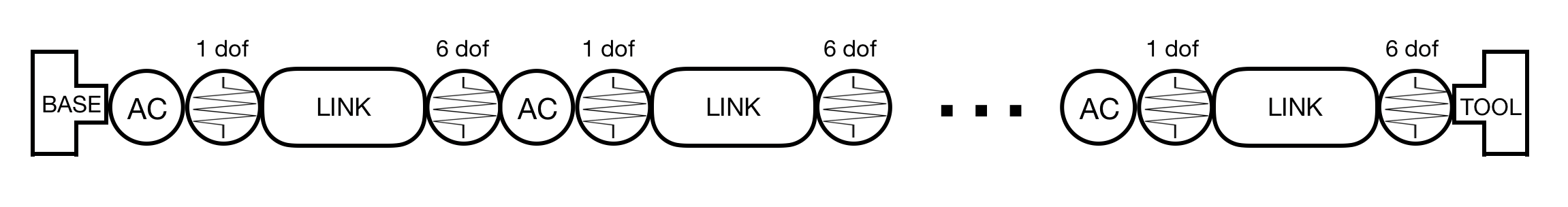}
    \caption{VJM model of Kuka iiwa 14 R820}
    \label{fig:vjm}
\end{figure}

Each elastic link is represented as a rigid link and a 6 dof virtual spring, similarly, each joint is represented as an actuated joint and a 1 dof virtual spring. The final VJM model is given by Figure \ref{fig:vjm} 

This VJM model correpond to the following extended model equation :
\begin{multline}
T_{ext}= T_{base}\cdot R_{z,q_{1}}\cdot R_{z,\theta_{1}} \cdot T_{z,d_{1}} \cdot H_{3D,\theta_{7}}\cdot R_{x,q_{2}}\cdot R_{x,\theta_{8}} \cdot T_{z,d_{2}} \cdot H_{3D,\theta_{14}} \cdot R_{z,q_{3}}\cdot R_{z,\theta_{15}}\cdot T_{z,d_{3}} \cdot H_{3D,\theta_{21}} \cdot R_{x,q_{4}}\cdot R_{x,\theta_{22}} \cdot T_{z,d_{4}} \cdot\\ H_{3D,\theta_{28}} \cdot R_{z,q_{5}}\cdot R_{z,\theta_{29}} \cdot T_{z,d_{5}} \cdot H_{3D,\theta_{35}} \cdot R_{x,q_{6}}\cdot R_{x,\theta_{36}} \cdot T_{z,d_{6}} \cdot H_{3D,\theta_{42}} \cdot R_{z,q_{7}}\cdot R_{z,\theta_{43}}\cdot T_{Tool}
\end{multline} 

The general equation between increments is as follows:
\begin{equation}
    \begin{bmatrix}
        0&J_\theta&J_q\\
        J_\theta^T&-K_\theta&0\\
        J_q^T&0&0\\
    \end{bmatrix}
    \cdot
    \begin{bmatrix}
        F\\\theta\\\Delta q\\
    \end{bmatrix}
    =
    \begin{bmatrix}
        \Delta t\\0\\0\\
    \end{bmatrix}
\end{equation}

The cartesian stiffness matrix has the final expression given by  equation \ref{eq_Kc}
\begin{equation}
    K_C = K^0_C-K^0_C\cdot J_q\cdot K_{Cq} \label{eq_Kc}
\end{equation}

where : \quad $ K^0_C = (J_\theta\cdot K_\theta^{-1}\cdot J_\theta^T)^{-1} $ \quad is the classical cartesian stiffness matrix\\

and: \quad
     $K_{Cq} = (J^T_q\cdot (K_C^0)^{-1}\cdot J_q)^{-1}\cdot J_q^T\cdot (K_C^0)^{-1},\, K_{C\theta} = K_\theta^{-1}\cdot J_\theta^T\cdot K_C $

\paragraph{MSA model:}
The basic expression for stiffness model is: \quad $W = K\cdot \Delta t$ \\ 
Using the cantilever beam representation of a link,  one can write the expression of the stiffness matrix as follow:
\begin{equation}
    K = 
\begin{bmatrix}
    \frac{E\cdot S}{L} & 0 & 0 & 0 & 0 & 0\\
    0 & \frac{12E\cdot I_z}{L^3} & 0 & 0 & 0 \frac{-6E\cdot I_z}{L^2}\\
    0&0&\frac{12E\cdot I_y}{L^3} & 0 & \frac{6E\cdot I_y}{L^2}&0\\
    0 &0 & 0 & \frac{G\cdot J}{L} & 0 & 0\\
    0&0&\frac{6E\cdot I_y}{L^2} & 0 & \frac{4E\cdot I_y}{L}& 0\\
    0& \frac{-6E\cdot I_z}{L^2} &0 & 0 & 0 & \frac{4E\cdot I_z}{L}\\
\end{bmatrix}
\end{equation}

We need to transform this model into a global coordinates system:
\begin{equation}
    \begin{bmatrix}
        K_{11}^{global} & K_{12}^{global}\\
        K_{21}^{global}& K_{22}^{global}\\
    \end{bmatrix}
    =
    \begin{bmatrix}
        R&0&0&0\\
        0&R&0&0\\
        0&0&R&0\\
        0&0&0&R\\
    \end{bmatrix}
    \begin{bmatrix}
        K_{11}^{local}&K_{12}^{local}\\
        K_{21}^{local}&K_{22}^{local}\\
    \end{bmatrix}
    \begin{bmatrix}
        R^T&0&0&0\\
        0&R^T&0&0\\
        0&0&R^T&0\\
        0&0&0&R^T\\
    \end{bmatrix}
\end{equation}

Then, the general expression of the wrenches applied in both sides of the link is:

\begin{equation}
    \begin{bmatrix}
        W_1\\W_2\\
    \end{bmatrix}
    =
    \begin{bmatrix}
        K_{11} & K_{12}\\
        K_{21}& K_{22}\\
    \end{bmatrix}_{12\times 12}
    \begin{bmatrix}
        \Delta t_1\\\Delta t_2\\
    \end{bmatrix}
\end{equation}
Where:
$$    K_{22} = K $$
$$    K_{11} = 
    \begin{bmatrix}
        R_z^\pi&0_{3\times 3}\\
        0_{3\times 3}&R_z^\pi\\
    \end{bmatrix}^T
    \cdot 
    K
    \cdot
    \begin{bmatrix}
        R_z^\pi&0_{3\times 3}\\
        0_{3\times 3}&R_z^\pi\\
    \end{bmatrix}$$
    $$     K_{12} = -
    \begin{bmatrix}
        I_{3 \times 3}& 0_{3 \times 3}\\
        [L\times]^T & I_{3 \times 3}\\
    \end{bmatrix}
    \cdot K_{22} $$
$$    K_{21} = -
    \begin{bmatrix}
        I_{3 \times 3} & 0_{3 \times 3}\\
        (L\times)&I_{3 \times 3}\\
    \end{bmatrix}
    \cdot K_{11}
$$

\begin{figure}[H]
    \centering
    \includegraphics[width=\linewidth]{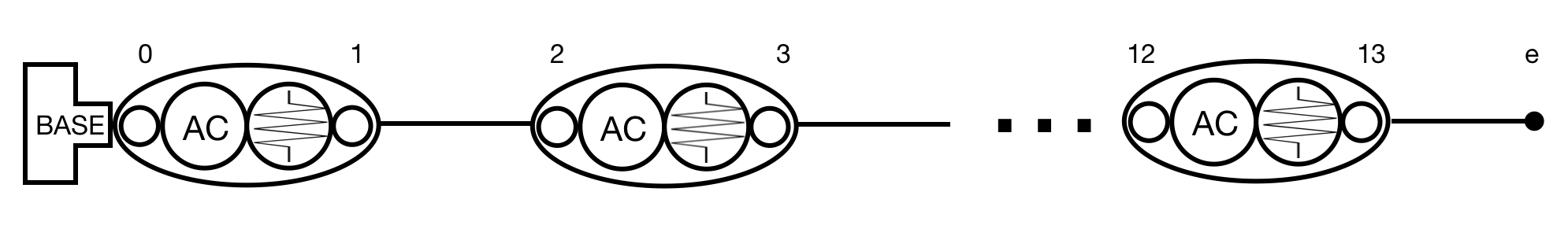}
    \caption{MSA model of Kuka iiwa 14 R820}
    \label{fig:msa}
\end{figure}

The MSA model of the robot is shown in Figure \ref{fig:msa}

\begin{equation}
    K_{links }= \left[ \begin{array}{cccc}
\begin{array}{cc} K_{11}^{1,2} & K_{12}^{1,2} \\ K_{21}^{1,2} & K_{22}^{1,2} \end{array} & \begin{array}{cc} 0 & 0 \\ 0 & 0 \end{array}& \begin{array}{cc} 0 & 0 \\ 0 & 0 \end{array} &\begin{array}{cc} 0 & 0 \\ 0 & 0 \end{array} 	  \\
\begin{array}{cc} 0 & 0 \\ 0 & 0 \end{array} &  \begin{array}{cc} K_{11}^{12} & K_{12}^{12} \\ K_{21}^{12} & K_{22}^{12} \end{array} &\begin{array}{cc} 0 & 0 \\ 0 & 0 \end{array} & \begin{array}{cc} 0 & 0 \\ 0 & 0 \end{array} \\
\begin{array}{cc} 0 & 0 \\ 0 & 0 \end{array} &\begin{array}{cc} 0 & 0 \\ 0 & 0 \end{array} &\begin{array}{cc} .... & .... \\ .... & .... \end{array} & \begin{array}{cc} 0 & 0 \\ 0 & 0 \end{array} \\
\begin{array}{cc} 0 & 0 \\ 0 & 0 \end{array} & \begin{array}{cc} 0 & 0 \\ 0 & 0 \end{array} & \begin{array}{cc} 0 & 0 \\ 0 & 0 \end{array} &  
\begin{array}{cc} K_{11}^{13, e} & K_{12}^{13, e} \\ K_{21}^{13, e} & K_{22}^{13, e} \end{array}
\end{array} \right]
\end{equation}

The details for each joint modeling is depicted below:
\begin{enumerate}

\item elastic support 0,1:
\begin{equation}
     \lambda _{*01}^{r} = \left[ \begin{array}{cccccc} 1&0&0&0&0&0 \\ 0&1&0&0&0&0 \\ 0&0&1&0&0&0 \\ 0&0&0&1&0&0 \\ 0&0&0&0&1&0 \end{array} \right],\,  \lambda _{*01}^{e} = \left[ \begin{array}{cccccc} 0&0&0&0&0&1  \end{array} \right]
\end{equation}
\begin{equation}
     \lambda _{*01}^{r} \cdot \Delta t_{1}= 0_{5\times 1}  \qquad \qquad \qquad 
K_{e} \cdot \lambda _{*01}^{e} \cdot \Delta t_{1} - \lambda _{*01}^{e} \cdot W_{1} = 0
\end{equation}

\item elastic joints with rotation about  x-axis  2,3  6,7  10,11:
\begin{equation}
    \lambda _{*ij}^{r} = \left[ \begin{array}{cccccc} 1&0&0&0&0&0 \\ 0&1&0&0&0&0 \\ 0&0&1&0&0&0 \\ 0&0&0&0&1&0 \\ 0&0&0&0&0&1 \end{array} \right],\,\lambda _{*ij}^{e} = \left[ \begin{array}{cccccc} 0&0&0&1&0&0  \end{array} \right]
\end{equation}

\begin{equation}
    \left[ \begin{array}{cccc} \lambda^{e}_{*ij} & 0_{6\times 6} & K^{e}_{ij}\cdot \lambda^{e}_{*ij} &  -K^{e}_{ij}\cdot \lambda^{e}_{*ij}  \\
I_{6\times 6 } & I_{6 \times 6} & 0_{6\times 6} & 0_{6\times 6} \\
0_{6\times 6} & 0_{6\times 6} & \lambda^{r}_{*ij} & -\lambda^{r}_{*ij}   \end{array} \right]
 \left[ \begin{array}{c} W_{i}\\ W_{j} \\ \Delta_{i} \\ \Delta_{j} \end{array} \right] = 
 \left[ \begin{array}{c} 0\\ 0 \\ 0  \end{array} \right]
\end{equation}

\item elastic joints with rotations about the z-axis  4,5  8,9  12,13:
\begin{equation}
    \lambda _{*ij}^{r} = \left[ \begin{array}{cccccc} 1&0&0&0&0&0 \\ 0&1&0&0&0&0 \\ 0&0&1&0&0&0 \\ 0&0&0&1&0&0 \\ 0&0&0&0&1&0 \end{array} \right],\,\lambda _{*ij}^{e} = \left[ \begin{array}{cccccc} 0&0&0&0&0&1  \end{array} \right]
\end{equation}

\begin{equation}
    \left[ \begin{array}{cccc} \lambda^{e}_{*ij} & 0_{6\times 6} & K^{e}_{ij}\cdot \lambda^{e}_{*ij} &  -K^{e}_{ij}\cdot \lambda^{e}_{*ij}  \\
I_{6\times 6 } & I_{6 \times 6} & 0_{6\times 6} & 0_{6\times 6} \\
0_{6\times 6} & 0_{6\times 6} & \lambda^{r}_{*ij} & -\lambda^{r}_{*ij}   \end{array} \right]
 \left[ \begin{array}{c} W_{i}\\ W_{j} \\ \Delta_{i} \\ \Delta_{j} \end{array} \right] = 
 \left[ \begin{array}{c} 0\\ 0 \\ 0  \end{array} \right]
\end{equation}

\item Aggregated model:
\begin{eqnarray} \label{aggre}
\left[ \begin{array}{cc} -I_{84\times 84} & K_{links} \\ 0_{35\times 84} & A_{agr} \\ B_{agr}& 0_{36\times 84} \\C_{agr}& D_{agr} \\ E_{agr} & 0_{6\times84} \end{array} \right]
 \left[ \begin{array}{c} W_{agr} \\ \Delta t_{agr}  \end{array} \right] = 
 \left[ \begin{array}{c} 0 _{168 \times 1} \\W_{e}  \end{array} \right]\\
 \end{eqnarray}

\begin{figure}[H]
    \centering
    \includegraphics[width=\linewidth]{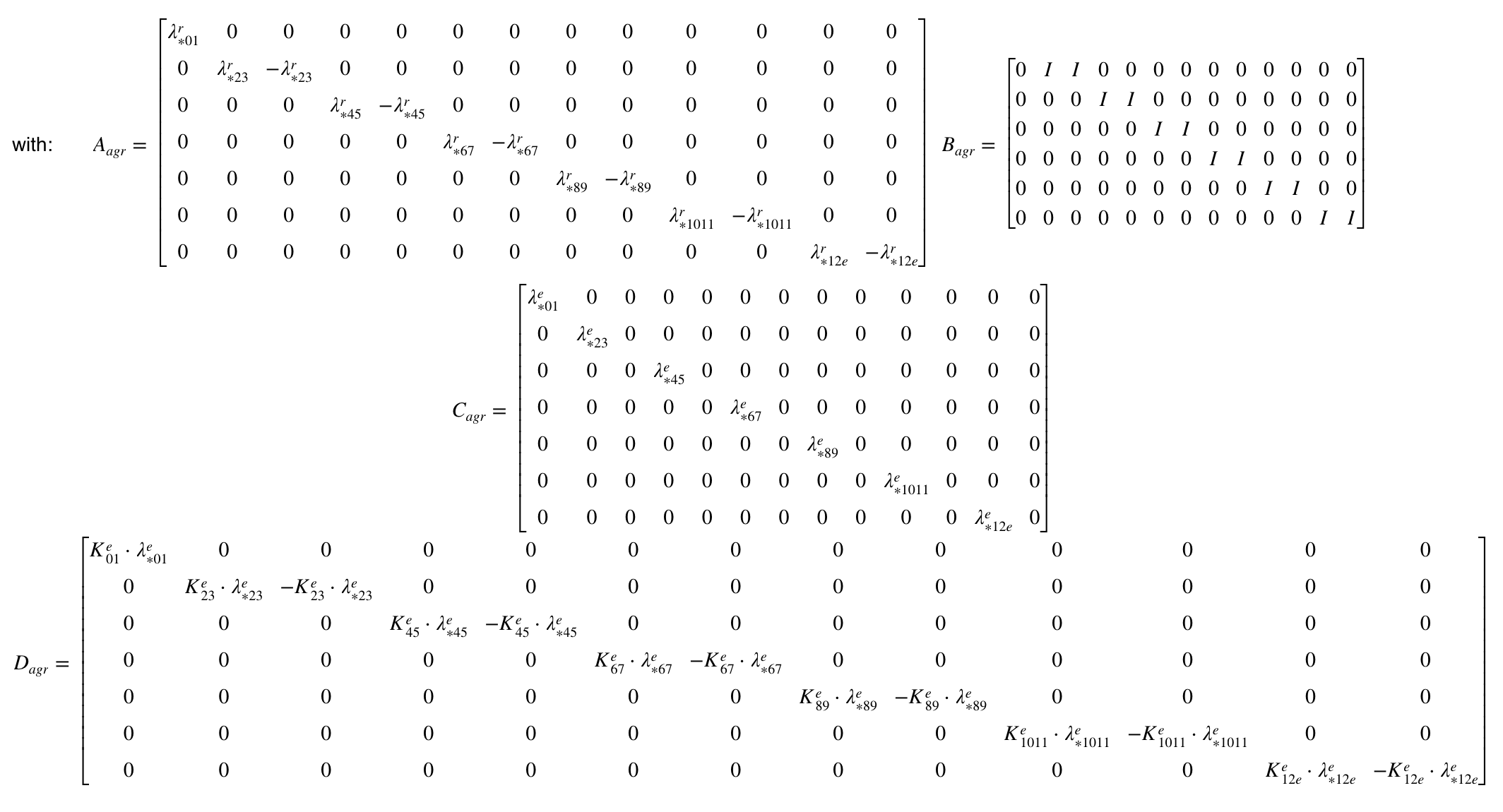}
\end{figure}

Equation \ref{aggre} can be written in the form:
\begin{equation}
 \begin{bmatrix}
     A&B\\
     C&D\\
 \end{bmatrix}
\cdot
\begin{bmatrix}
    \mu\\\Delta t_e\\
\end{bmatrix}
=
\begin{bmatrix}
    b\\W_e\\
\end{bmatrix}\\
\end{equation}
And the final expression of $K_c$ is given by:
$$
K_C = D-C\cdot A^{-1}\cdot B
$$
\end{enumerate}

\subsection{Geometric Calibration:}
The basic equations for the identification using full pose measurements is 
\begin{equation}
    t= g(q, \pi)
\end{equation}
where:\quad $ t_i = (p_{xi}, p_{yi}, p_{zi}, \phi_{xi}, \phi_{yi}, \phi_{zi})^T$ is six dimensional location vector, $g$ is the manipulator extended geometric model, $q$ is the vector of actuated coordinates and $\pi  = \pi + \Delta \pi $ is the vector of parameters.\\
The goal is the identification of robot model parameters using direct measurement only without using orientation components, the basic equation for identification is:
\begin{equation}
    \sum _{i=1}^{m} \sum_{j=1}^{n} || p_{ij} - g_{0ij}^{(p)} - J_{\pi ij}^{(p)} \Delta \pi ||^2 \longrightarrow min
\end{equation}

The geometric model obtained via homogeneous transformations can be presented by the matrix product:
\begin{equation}
    T_i^j = T_{base} T_{robot}(q_i, \pi) T_{tool}^j
\end{equation}
The cartesian coordinates of reference points $p_i^j, j=1, ...n$ corresponding to the configuration $q_i$ can be expressed in the following form:

\begin{equation}
    p_i^j= p_{base} +R_{base} p_{robot}(q_i, \pi) +R_{base}R_{robot} (q_i, \pi) p_{tool}^j
\end{equation}

Where: $p_{base}, R_{base}, p_{tool}^j, \pi$ are unknown parameters  \\
The procedure of identification is divided into two steps \cite{klimchik2014geometric}:
\begin{enumerate}
    \item \textbf{Step1}: parameters identification  of tool and base transformations  $p_{base}, R_{base}, p_{tool}^j$
    \begin{equation}
        \left[ \begin{array}{cccc} p_{base} & r_{base} & u_{tool},...& u_{tool}^n \end{array} \right]
    =( \sum_{i=1}^{m} A_i^{jT} A_i^{j})^{-1}  (\sum_{i=1}^{m} A_i^{jT} \Delta \pi)
    \end{equation}
      
    Where: $A_i^j=\left[ \begin{array}{cccccc} I & [\Tilde{p_{base}}]^T  & R_{robot}^i & 0 &.
    ....& 0 \\ I & [\Tilde{p_{base}}]^T  & 0 & R_{robot}^i &.....& 0 \\  .... & ....  & .... & .... &....& ....  \\
    I & [\Tilde{p_{base}}]^T & 0 & 0 & ....& R_{robot}^i\end{array} \right] $
    
    and $p_{tool}^j = R_{base}^T \cdot u_{tool}^j$

    \item \textbf{Step2:} Identification of the   Electrostatic and geometric parameters of the manipulator $\pi$ \\
    Basic equation of identification : \quad $ p_i^j = p_{robot}^i + J_{\pi i}^{j(p)} \pi  $, where $\pi$ are the unknown parameters to be identified \\
    The solution of the identification problem is given by 
    \begin{equation}
        \pi = ( \sum_{i=1}^{m} J_i^{j(p)T} J_i^{j(p)})^{-1}  (\sum_{i=1}^{m} J_i^{j((p)T} \Delta \pi_i^j)
    \end{equation}
\end{enumerate}

\subsection{Modeling:}
To perform the identification task, one need to develop a suitable geometric model, which properly describe the relation between the manipulator geometric parameters (link length and joint angles) and the end effector location (position and orientation), for this, we construct the complete and obviously irreducible model in the form of homogeneous matrices product
\begin{itemize}
    \item The base transformation is $T_{base}= [T_x T_y T_z R_x R_y R_z] $
    \item The joint and link transformations (for revolute joint) $T_{joint, j}  T_{link, j}= R_{e,j}(q_j, \pi_{qj})[T_uT_vR_uR_v]$
    Where $e_j$ is the joint axis and $u_j$ and $v_j$ are the axes orthogonal to $e_j$
    \item The tool transformation $T_{tool}= [T_x T_y T_z R_x R_y R_z] $\\
    
The next step is the elimination of non identifiable and semi identifiable parameters in accordance with specific rules for different nature and structure of consecutive joints 
    \item In the case of consecutive revolute joints $R_{e,j}(q_j, \pi_{qj})$
    \begin{itemize}
        \item if $e_j \perp e_{j-1}$ eliminate the term $R_{u, L_{j-1}}$ or $R_{v, L_{j-1}}$ that correspond to $R_{e,j}$
        \item if $e_j \parallel e_{j-1}$ eliminate the term $T_{u, L_{j-k}}$ or $T_{v, L_{j-k}}$ that define the translation orthogonal to the joint axes for which k is minimum $(k\geq 1)$
    \end{itemize}
\end{itemize} 

\subsection{Design of calibration experiments:}

In the case of serial manipulator with revolute joints, the expression of end effector position is computed using the following formula:
\begin{equation} \label{eq1}
x^k= \sum _{i=0}^{n} (l_i^0 + \Delta l_i) \cos{(\theta_i^0 + \Delta \theta_i)} \qquad 
y^k= \sum _{i=0}^{n} (l_i^0 + \Delta l_i) \sin{(\theta_i^0 + \Delta \theta_i)}
\end{equation}
Where: 
$l_i^0$ are the nominal links lengths and $\Delta l_i$ their deviations, $q_j^0$ are nominal joints coordinates, $\theta_i$ are defined as $ \theta_i= \sum _{k=1}^{i}$ $ q_k^0 $ and   $ \Delta \theta_i= \sum _{k=1}^{i} \Delta q_k^0 $ are the joints offsets\\

To take into account the impact of measurement noise, the calibration equation derived from \ref{eq1} becomes:

\begin{equation} \label{eq2}
x^k= \sum _{i=0}^{n} (l_i^0 + \Delta \theta_i) \cos{(\theta_i^0 + \Delta \theta_i)} + \epsilon_x^k \qquad 
y^k= \sum _{i=0}^{n} (l_i^0 + \Delta \theta_i) \sin{(\theta_i^0 + \Delta \theta_i)}  +  \epsilon_y^k
\end{equation}

To find the desired parameters using the noise corrupted measurements, the least square technique is applied, this approach aims at minimizing the square sum of the residuals in \ref{eq2} simultaneously 

\begin{equation} 
\sum _{k=1}^{m} \left( \sum _{i=0}^{n} (l_i^0 + \Delta l_i) \cos{(\theta_i^0 + \Delta \theta_i)} - x_k\right)^2 +  
\sum _{k=1}^{m} \left( \sum _{i=0}^{n} (l_i^0 + \Delta l_i) \sin{(\theta_i^0 + \Delta \theta_i)} - y_k \right) ^2 \longrightarrow min
\end{equation}

Collecting the unknown parameters $\Delta l_i $ and $\Delta \theta_i$ into the vector $\Delta \pi$ and the measurements into the vector $\Delta P^k$  equation \ref{eq1} can be rewritten as:
\begin{equation}
    \Delta P^k= J^k \Delta \pi
\end{equation}

Where $J$ is the jacobian matrix, then one can get the unknown parameters using the least squares technique that leads to 

\begin{equation}\label{eq4}
\Delta \pi = \left( \sum _{k=1}^{m} J^{kT} J^k \right)^{-1}   \sum _{k=1}^{m} J^{kT} \Delta P^k
\end{equation}
Where the subscript k indicate the experiment number and m is the number of measurements 
Considering that each measurement is corrupted by an unbiased random Gaussian noise with standard deviation $\sigma$, the identification  accuracy of the parameters $\Delta \pi$ can be evaluated via the covariance matrix, which is computed as follow:
\begin{equation}
    cov (\Delta \pi)= \sigma^2 (\sum ^m _{k=1}  J^{kT} J^k )^{-1}
\end{equation}
With this expression, it is possible to choose the measurement configuration that yield parameters less sensitive to measurement noise, this procedure is referred to as design of calibration experiments, the optimality condition for the calibration plan proposed in \cite{klimchik_antropomorphic} where one need to ensure that the information matrix is diagonal yields to the D-optimal plan of experiments that is satisfied when:
\begin{equation} \label{eq3}
\begin{array}{c}
\sum_{k=1}^m \cos (\sum_{s=1}^i q_s^k- \sum_{k=1}^j q_s^k ), \quad \forall i>j\\
\sum_{k=1}^m \sin (\sum_{s=1}^i q_s^k- \sum_{k=1}^j q_s^k ), \quad \forall i>j      
\end{array}    
\end{equation}
The above presented equations define the desired set of optimal measurements configurations.

\subsection{Geometrical patterns for measurement pose selection:}
In the following we are going to introduce some important properties of the optimality condition \ref{eq3} that allow us to reduce the problem complexity \cite{klimchik2014geometrical}

\begin{enumerate}
    \item \textit{Superposition of optimal plans also gives an optimal plan for this} ; this is due to the additivity of the the operations included in \ref{eq3}.Using this property , it is possible to generate optimal plan with a large number of measurements configurations using simple sets.
    \item \textit{The angles $q_1, q_2, ...., q_n$} can be rearranged in the optimal plan in an arbitrary way without loss of the optimality condition \ref{eq3}
    \item \textit{Optimal plan for n-link manipulator can be obtained using two lower-order optimal plans for n1- and n2-link manipulators, where $n_1 + n_2 = n+1$} , This property gives an elegant technique to generate optimal plan of calibration experiments without straightforward solution of the system \ref{eq3} 
\end{enumerate}

the following present geometrical patterns for typical serial manipulators that can be used to generate optimal plans. In the frame of these patterns, all variables $\alpha_i \beta_i$ and $\gamma $ and $\delta$ are treated as arbitrary angles:

For n = 3, m = 3, the geometrical pattern can be presented as
\begin{eqnarray} \begin{array}{ccc}
q_1^1 = \alpha_1 & q_2^1 = \beta  & q_3^1 = \gamma  \\
q_1^2 = \alpha_2 & q_2^2 = \beta + 2\pi/3 & q_3^2  = \gamma  + 2\pi/3 \\
q_1^3 = \alpha_3 & q_2^3 = \beta - 2\pi/3 & q_3^3  =\gamma  - 2\pi/3
\end{array}
\end{eqnarray} 

For n = 4, m = 4, the geometrical pattern can be presented as
\begin{eqnarray} \begin{array}{cccc} \label{eq6}
q_1^1 = \alpha_1 & q_2^1 = \beta_1  & q_3^1 = \gamma & q_4^1 = \delta \\
q_1^2 = \alpha_2 & q_2^2 = \beta_1 + \pi  & q_3^2  = \gamma & q_4^2= \delta + \pi\\
q_1^3 = \alpha_3 & q_2^3 = \beta_2  & q_3^3  =\gamma  + \pi & q_4^3  =\delta  + \beta_1 - \beta_2  \\
q_1^4 = \alpha_4 & q_2^4 = \beta_2 + \pi  & q_3^4 =\gamma  + \pi& q_4^3  =\delta  + \beta_1 - \beta_2 + \pi\\ 
\end{array}
\end{eqnarray}

\begin{figure}[H]
    \centering
    \includegraphics[width=\textwidth]{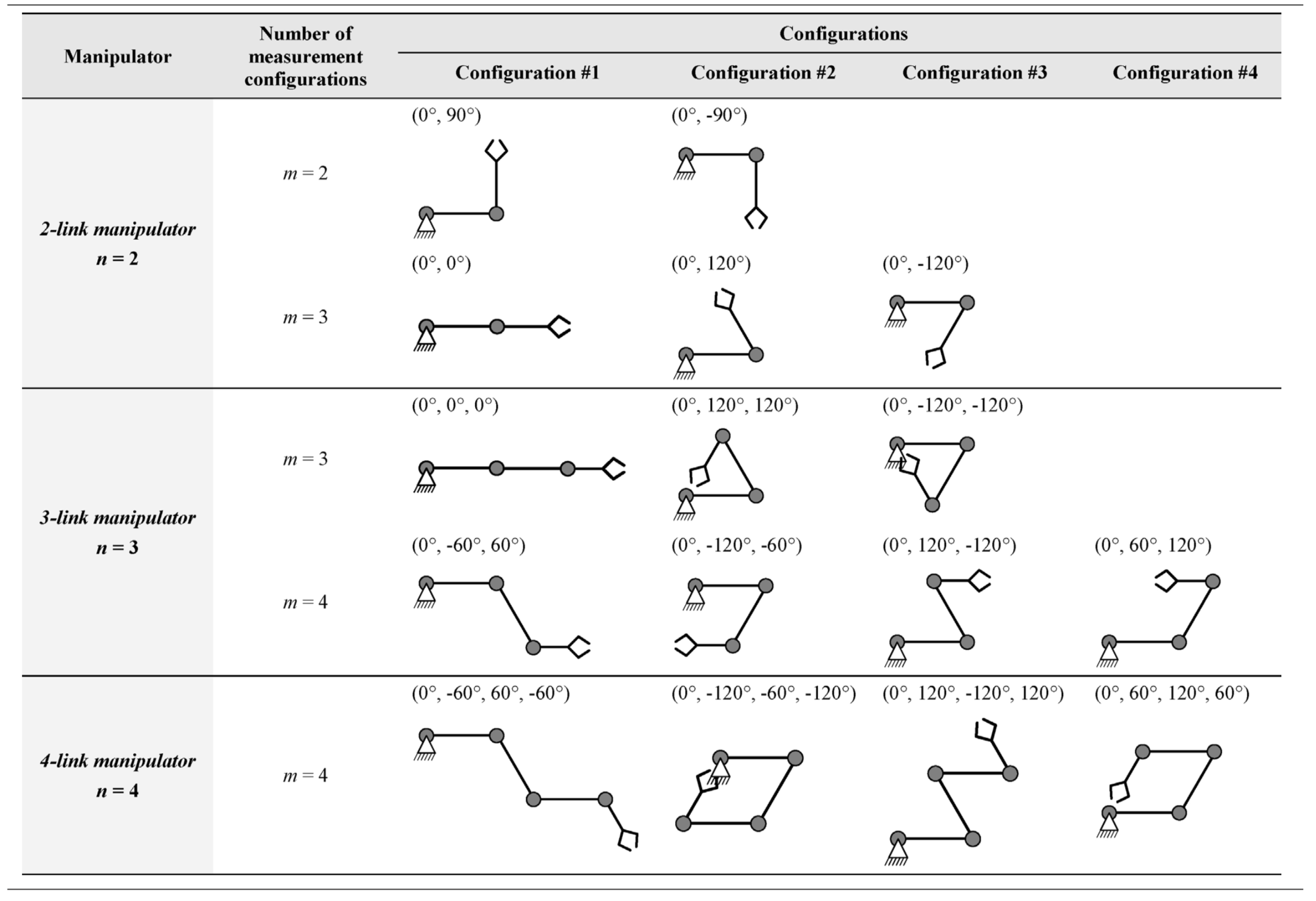}
    \caption{Optimal measurement configurations for typical planar manipulators \cite{klimchik_math}}
    \label{planar}
\end{figure}

\subsection{Extension to the 3D case:}

The simplest way to extend the proposed “rule of thumb” to the 3D case is to apply the following procedure:\\
Step (a): decompose the spatial manipulator into a set of planar serial sub-chains;\\
Step (b): apply the developed rule to each sub-chain separately,without assigning certain values to the joint coordinates that can be selected arbitrary;\\
Step (c): aggregate the obtained sub-chain joint coordinates in order to find configurations of the entire manipulator (where some values are still arbitrary);\\
Step (d): apply the developed rule to the each set of the arbitrary coordinates, i.e. ensuring that the sums of sines and cosines are equal to zero for all of them.

\section{Case study: calibration of the 7 dof manipulator KUKA IIWA}
\subsection{Calibration experiment:}
\begin{figure}[H]
    \centering
    \includegraphics[width=0.8\textwidth]{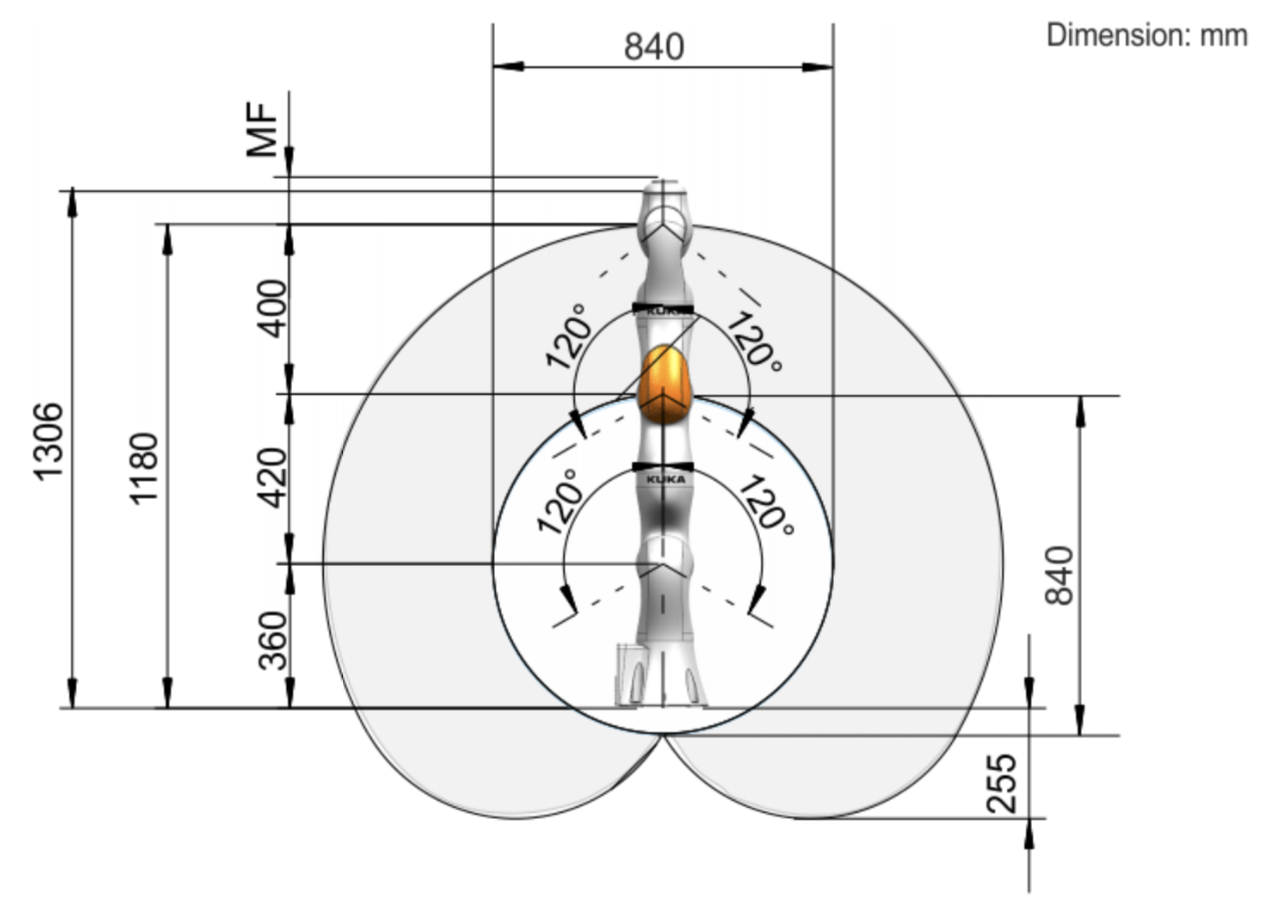}
    \caption{LBR iiwa 14 R820 working envelope, side view}
    \label{fig_kuka}
\end{figure}

Let us consider the 7 dof. serial manipulator with seven revolute joints and six links, Kuka IIWA 14 R820 depicted in Figure \ref{fig_kuka}  

To complete the identification task, the irreducible model in the form of homogeneous matrix product as described before was build for Kuka iiWA: 

\begin{align*}
    T &= [T_x T_y T_z R_x R_y R_z]_b R_z(q_1+\cancel{\Delta q_1})[T_x T_y \cancel{R_x} R_y] R_x (q_2 + \cancel{\Delta q_2})[T_y T_z R_y \cancel{R_z}]R_z(q_3 + \Delta q_3)\\&[T_x T_y \cancel{R_x}Ry]R_x(q_4+\Delta q_4)[T_y T_z R_y \cancel{R_z}]R_z(q_5 + \Delta q_5)[T_x T_y \cancel{R_x} R_y]R_x(q_6 + \Delta q_6)[T_y T_z R_y \cancel{R_z}]\\&R_z(q_7 + \cancel{\Delta q_7})[\cancel{T_x}\cancel{T_y} \cancel{R_x} \cancel{R_z}][\cancel{T_x}\cancel{T_y}\cancel{T_z}\cancel{R_x}\cancel{R_y}\cancel{R_z}]\\
       T_{robot} &= R_z(q_1)\cdot T_x(\Delta l_{1x})  \cdot T_y(\Delta l_{1y})\cdot R_y(\Delta q_{1y})\cdot R_x (q_2 + \Delta q_2)\cdot T_y(\Delta l_{2y})\cdot T_z(d_1 + \Delta l_{2z})\cdot R_y(\Delta q_{2y}) \cdot \\ & R_z(q_3 + \Delta q_3) \cdot T_x(\Delta l_{3x})\cdot  T_y(\Delta l_{3y}) \cdot Ry(\Delta q_{3y}) \cdot R_x(q_4+\Delta q_4) \cdot T_y(\Delta l_{4y})\cdot  T_z(d_2 + \Delta l_{4z})\cdot R_y(\Delta q_{4y}) \cdot  \\ & R_z(q_5 + \Delta q_5) \cdot T_x(\Delta l_{5x})\cdot T_y(\Delta l_{5y}) \cdot R_y(\Delta q_{5y}) \cdot R_x(q_6 + \Delta q_6) \cdot T_y(\Delta l_{6y})\cdot T_z(\Delta l_{6z})\cdot R_y(\Delta q_{6y}) \cdot  R_z(q_7)
    \end{align*}

\begin{figure}[H]
    \centering
    \includegraphics[width=0.4\textwidth]{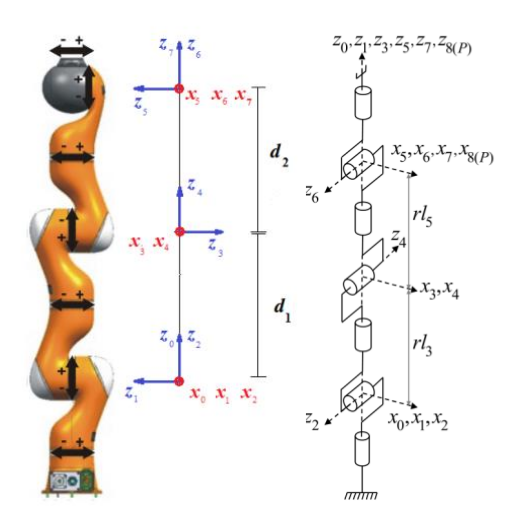}
    \caption{Representation of the 7dof manipulator KUKA iiwa}
    \label{DH}
\end{figure}

\begin{figure}[H]
    \centering
    \includegraphics[width=0.4\textwidth]{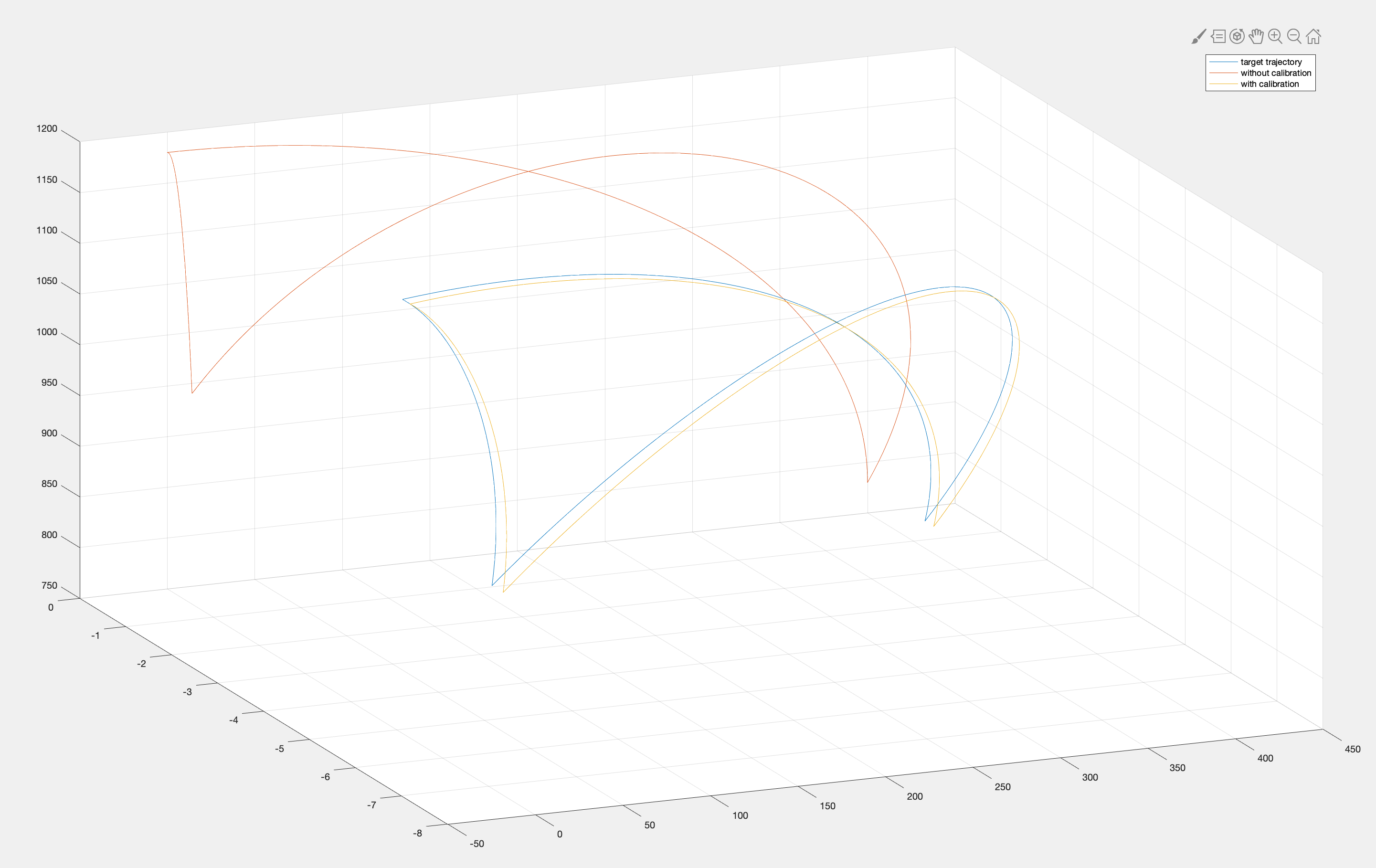}
    \includegraphics[width=0.4\textwidth]{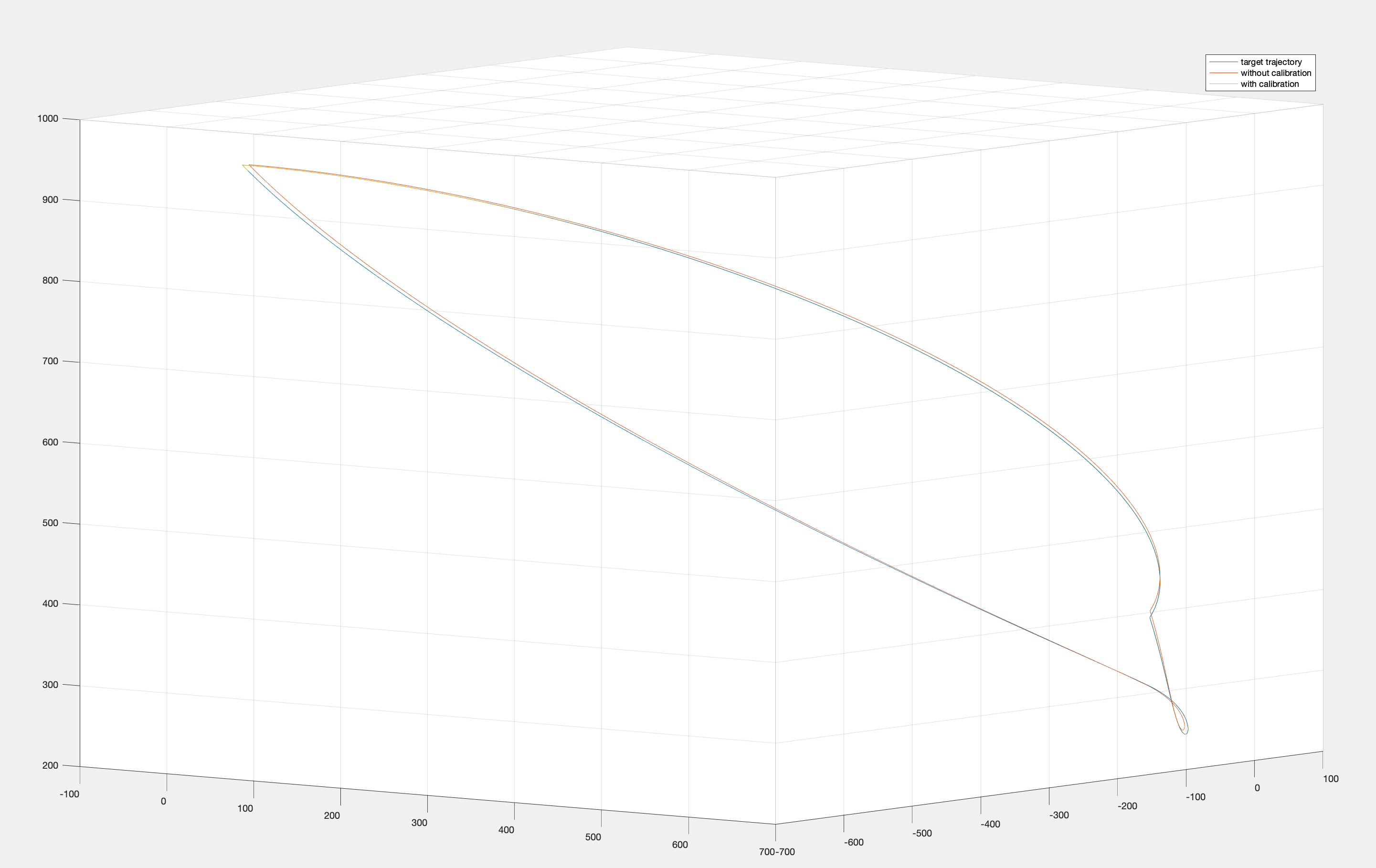}
    \includegraphics[width=0.4\textwidth]{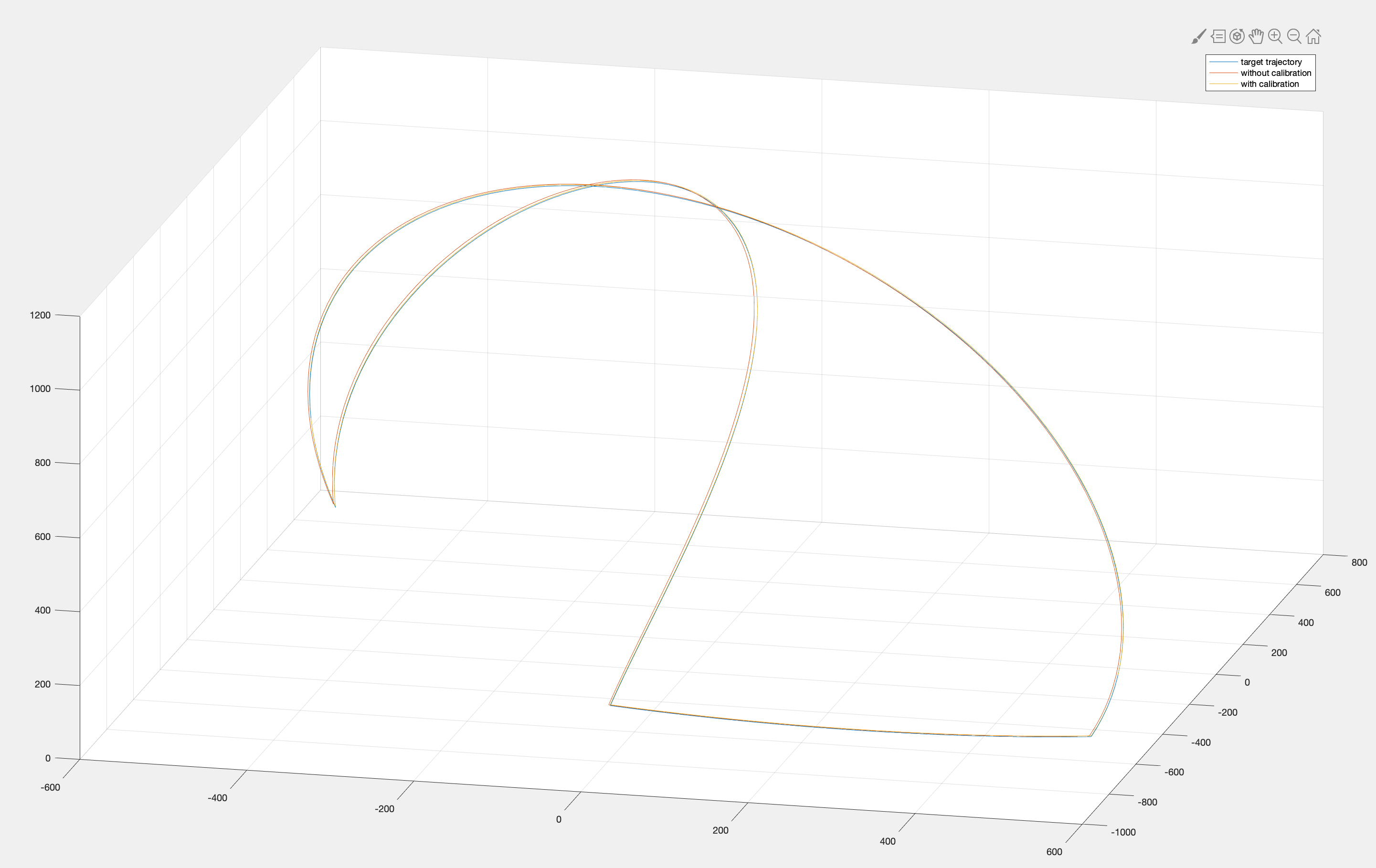}
    \caption{Trajectories obtained with and without calibration (target trajectory(blue), trajectory without calibration (red), trajectory with calibration (yellow)}
    \label{Figure4}
\end{figure}

To simulate the calibration procedure, we take ideal parameters for which we add some random noise and  we compute the  calibration procedure described before using a \textit{Matlab} code, Figure \ref{Figure4} shows the results obtained for three different trajectories (trajectory obtained with and without calibration).

We can clearly see the trajectory with calibration follow better the desired trajectory in all three cases, especially in the last two pictures where the trajectories are almost colinear.

Table \ref{table1} shows the results of the parameters identification with and without calibration, the parameters obtained after calibration are more accurate and closer to the real ones.

\begin{table}[H]
    \centering
    \begin{tabular}{>{\centering}p{2cm} >{\centering}p{2cm} >{\centering}p{2cm} >{\centering}p{2cm} >{\centering}p{2cm}}
        \multicolumn{5}{c}{\textbf{Parameters Identification}} \\
        \hline
         \textbf{Parameter}&  \textbf{Real value} & \textbf{Calibration} &\textbf{No Calibration} & \textbf{Improvement Factor}\tabularnewline
         \hline
         $p_{x1}$&-0.0051& -0.0088&0.0000&1.37\tabularnewline
         $p_{y1}$ & -0.0023& 0.0187&0.0000&1.10\tabularnewline
         $\phi_{y1}$ & -0.0049 & -0.0049&0.0000&236.95\tabularnewline
         $\Delta q_2$& 0.0089 & 0.0088&0.0000&178.35\tabularnewline
         $p_{y2}$& -0.0058 & -0.0380&0.0000&0.17\tabularnewline
         $p_{2z}$& 423.8028 & 423.7902&420.0000&301.81\tabularnewline
         $\phi_{y2}$& -0.0023& -0.0024&0.0000&56.69 \tabularnewline
         $\Delta q_3$& -0.0058& -0.0058&0.0000&500.43\tabularnewline
         $p_{x3}$& 0.0074& -0.0046&0.0000&0.61\tabularnewline
         $p_{y3}$& -0.0097& -0.0132&0.0000&2.82\tabularnewline
         $\phi_{y3}$& -0.0052& -0.0052&0.0000&76.20\tabularnewline
         $\Delta q_4$& 0.0036& 0.0035&0.0000&45.79\tabularnewline
         $p_{y4}$& 0.0035& -0.0196&0.0000&0.15\tabularnewline
         $p_{z4}$& 399.3576& 399.3678&400.0000&62.98\tabularnewline
         $\phi_{y4}$& -0.0050& -0.0050&0.0000&127.71\tabularnewline
         $\Delta q_5$& 0.0063& 0.0064&0.0000&62.00\tabularnewline
         $p_{x5}$& -0.0046& -0.0283&0.0000&0.18\tabularnewline
         $p_{y5}$& -0.0057& -0.0111&0.0000&1.03\tabularnewline
         $\phi_{y5}$& $6.762e^{-4}$& 0.0007&0.0000&12.81\tabularnewline
         $\Delta q_6$& 0.0023& 0.0024&0.0000&22.00\tabularnewline
         $p_{y6}$& 0.0048& 0.0096&0.0000&1.00\tabularnewline
         $p_{z6}$& 0.0048& 0.6044&0.0000&0.00\tabularnewline
         $\phi_{y6}$& -0.0041& -0.0041&0.0000&201.38\tabularnewline
         base x & 0.0000 & -0.0098 & 0.0000& 0.00\tabularnewline
         base y& 0.0000 & -0.0192& 0.0000 & 0.00 \tabularnewline
         base z& 365.5582 & 365.5518&360.0000&537.78\tabularnewline
         tool 1x&0.0000&0.0030&0.0000&0.00\tabularnewline
         tool 1y&0.0000&-0.0111&0.0000&0.00\tabularnewline
         tool 1z&89.6944&89.0951&90.0000&0.50\tabularnewline
         tool 2x&0.0000&-0.0014&0.0000&0.00\tabularnewline
         tool 2y&-77.5044&-77.5058&-77.9423&336.76\tabularnewline
         tool 2z&-44.7472&-45.3375&-45.0000&0.42\tabularnewline
         tool 3x&0.0000&0.0204&0.0000&0.00\tabularnewline
         tool 3y&78.6892&78.6877&77.9423&497.93\tabularnewline
         tool 3z&-45.4312&-46.0641&-45.0000&0.68\tabularnewline
         \hline
         \multicolumn{4}{c}{Average: } & 93.36\tabularnewline
    \end{tabular}
    \caption{Results of the parameters identification with and without calibration}
    \label{table1}
\end{table}

\subsection{Calibration using design of experiments:}

In order to find the optimal pose configuration for the manipulator, the first thing to do is to decompose the spatial manipulator into a set of planar kinematic serial sub-chains, we define each joint of the manipulator as $q_n$ where \textit{n} denotes the joint number of the robot. According to the geometrical pattern presented above, and by studying the robot structure, we naturally divide the manipulator into the two following planar sub-chains:
\begin{enumerate}
    \item $(q_1, q_3, q_5, q_7)$
    \item $(q_{virtual}, q_2, q_4, q_6)$
\end{enumerate}

Note that, because the first angle doesn't appear in the system \ref{eq3} (the calibration plan is invariant with respect to the first joint), we added a virtual joint at the beginning of the second sub-chain. 
Using this decomposition, the property 3 described above and the results of previous works  shown in  Figure \ref{planar} \cite{klimchik_math} corresponding to the 4-link manipulator with $m=4$ (equation \ref{eq6}), we built 16 configurations that reflect all the possible poses obtained when combining the two sub-chains. (table\ref{table2})

\begin{table}[H]
    \centering
    \begin{tabular}{ccccccc}
     $\boldsymbol{q_1}$ & $\boldsymbol{q_2}$ &$\boldsymbol{q_3}$ &$\boldsymbol{q_4}$ & $\boldsymbol{q_5}$ & $\boldsymbol{q_6}$ & $\boldsymbol{q_7}$ \\
     \hline
     $\alpha_1$ & $\epsilon_1$ & $\beta_1$ & $\chi$ & $\gamma$ & $\phi$ & $\delta$ \\
     $\alpha_1$ & $\epsilon_1+\pi$ & $\beta_1$ & $\chi$ & $\gamma$ & $\phi+\pi$ & $\delta$ \\
     $\alpha_1$ & $\epsilon_2$ & $\beta_1$ & $\chi+\pi$ & $\gamma$ & $\phi+\epsilon_1-\epsilon_2$ & $\delta$ \\
     $\alpha_1$ & $\epsilon_2+\pi$ & $\beta_1$ & $\chi+\pi$ & $\gamma$ & $\phi+\epsilon_1-\epsilon_2+\pi$ & $\delta$ \\
     $\alpha_2$ & $\epsilon_1$ & $\beta_1+\pi$ & $\chi$ & $\gamma$ & $\phi$ & $\delta+\pi$ \\
     $\alpha_2$ & $\epsilon_1+\pi$ & $\beta_1+\pi$ & $\chi$ & $\gamma$ & $\phi+\pi$ & $\delta+\pi$ \\
     $\alpha_2$ & $\epsilon_2$ & $\beta_1+\pi$ & $\chi+\pi$ & $\gamma$ & $\phi+\epsilon_1-\epsilon_2$ & $\delta+\pi$ \\
     $\alpha_2$ & $\epsilon_2+\pi$ & $\beta_1+\pi$ & $\chi+\pi$ & $\gamma$ & $\phi+\epsilon_1-\epsilon_2+\pi$ & $\delta+\pi$ \\
     $\alpha_3$ & $\epsilon_1$ & $\beta_2$ & $\chi$ & $\gamma+\pi$ & $\phi$ & $\delta+\beta_1-\beta_2$ \\
     $\alpha_3$ & $\epsilon_1+\pi$ & $\beta_2$ & $\chi$ & $\gamma+\pi$ & $\phi+\pi$ & $\delta+\beta_1-\beta_2$ \\
     $\alpha_3$ & $\epsilon_2$ & $\beta_2$ & $\chi+\pi$ & $\gamma+\pi$ & $\phi+\epsilon_1-\epsilon_2$ & $\delta+\beta_1-\beta_2$ \\
     $\alpha_3$ & $\epsilon_2+\pi$ & $\beta_2$ & $\chi+\pi$ & $\gamma+\pi$ & $\phi+\epsilon_1-\epsilon_2+\pi$ & $\delta+\beta_1-\beta_2$ \\
     $\alpha_4$ & $\epsilon_1$ & $\beta_2+\pi$ & $\chi$ & $\gamma+\pi$ & $\phi$ & $\delta+\beta_1-\beta_2+\pi$ \\
     $\alpha_4$ & $\epsilon_1+\pi$ & $\beta_2+\pi$ & $\chi$ & $\gamma+\pi$ & $\phi+\pi$ & $\delta+\beta_1-\beta_2+\pi$ \\
     $\alpha_4$ & $\epsilon_2$ & $\beta_2+\pi$ & $\chi+\pi$ & $\gamma+\pi$ & $\phi+\epsilon_1-\epsilon_2$ & $\delta+\beta_1-\beta_2+\pi$ \\
     $\alpha_4$ & $\epsilon_2+\pi$ & $\beta_2+\pi$ & $\chi+\pi$ & $\gamma+\pi$ & $\phi+\epsilon_1-\epsilon_2+\pi$ & $\delta+\beta_1-\beta_2+\pi$ \\
    \end{tabular}
    \caption{Optimal Plan for the entire manipulator with 16 measurements configurations}
    \label{table2}
\end{table}

Also,  to build the final optimal plan, a \textit{Matlab} code  was developed to set the still remaining arbitrary angles that satisfy the optimality condition \ref{eq3} for the entire manipulator. Taking into account the joint limits, the arbitrary angles have been set to: $\alpha_1 =0^\circ $, $\alpha_2 =0^\circ $, $\alpha_3 =0^\circ $, $\alpha_4 =0^\circ $, $\epsilon_1 =-90^\circ $, $\epsilon_2 =-90^\circ $, $\beta_1 = -90^\circ $, $\beta_2 =-130^\circ $, $\chi = -110^\circ $, $\gamma =-160^\circ $, $\phi = -110^\circ$, $\delta =-170^\circ $, and the Table \ref{configs} was built which shows 16 optimal configurations for our robot. One point that one should noted is that, because the calibration plan is invariant with respect to the first joint like stated before, we can arbitrarily assign joint $q_1$ to make the end effector facing the direction where the laser tracker should be positioned when performing the experiment.

\begin{table}[H]
\centering
\begin{tabular}{c c c c c c c c}
\multicolumn{8}{c}{\textbf{Optimal Pose Measurements}} \\
\hline
\textbf{Configuration} & \textbf{Joint 1} & \textbf{Joint 2} & \textbf{Joint 3} & \textbf{Joint 4} & \textbf{Joint 5} & \textbf{Joint 6} & \textbf{Joint 7} \\ \hline
1 & $-20^\circ$ & $-90^\circ$ & $-110^\circ$ & $-160^\circ$ & $-110^\circ$ & $-110^\circ$ & $-170^\circ$ \\ 
2 & $40^\circ$ & $90^\circ$ & $-90^\circ$ & $-110^\circ$ & $-160^\circ$ & $70^\circ$ & $-170^\circ$ \\ 
3 & $135^\circ$ & $-90^\circ$ & $-90^\circ$ & $70^\circ$ & $-160^\circ$ & $-110^\circ$ & $-170^\circ$ \\ 
4 & $-140^\circ$ & $90^\circ$ & $-90^\circ$ & $70^\circ$ & $-160^\circ$ & $70^\circ$ & $-170^\circ$ \\ 
5 & $20^\circ$ & $-90^\circ$ & $90^\circ$ & $-110^\circ$ & $-160^\circ$ & $-110^\circ$ & $10^\circ$ \\ 
6 & $-45^\circ$ & $90^\circ$ & $90^\circ$ & $-110^\circ$ & $-160^\circ$ & $70^\circ$ & $10^\circ$ \\ 
7 & $-140^\circ$ & $-90^\circ$ & $90^\circ$ & $70^\circ$ & $-160^\circ$ & $-110^\circ$ & $10^\circ$ \\ 
8 & $-140^\circ$ & $90^\circ$ & $90^\circ$ & $70^\circ$ & $-160^\circ$ & $70^\circ$ & $10^\circ$ \\ 
9 & $-165^\circ$ & $-90^\circ$ & $-130^\circ$ & $-110^\circ$ & $20^\circ$ & $-110^\circ$ & $-130^\circ$ \\
10 & $165^\circ$ & $90^\circ$ & $-130^\circ$ &$ -110^\circ$ & $20^\circ$ & $70^\circ$ & $-130^\circ$ \\ 
11 & $0^\circ$ & $-90^\circ$ & $-130^\circ$ & $70^\circ$ & $20^\circ$ & $-110^\circ$ & $-130^\circ$ \\ 
12 & $10^\circ$ & $90^\circ$& $-130^\circ$ & $70^\circ$ & $20^\circ$ & $70^\circ$ & $-130^\circ$ \\ 
13 & $165^\circ$ & $-90^\circ$& $50^\circ$ & $-110^\circ$ & $20^\circ$ & $-110^\circ$ & $50^\circ$ \\ 
14 & $-160^\circ$ & $90^\circ$& $50^\circ$ & $-110^\circ$ & $20^\circ$ & $70^\circ$ & $50^\circ$ \\
15 & $0^\circ$ & $-90^\circ$& $50^\circ$ & $70^\circ$ & $20^\circ$ & $-110^\circ$ & $50^\circ$ \\
16 & $-15^\circ$ & $90^\circ$& $50^\circ$ & $70^\circ$ & $20^\circ$ & $70^\circ$ & $50^\circ$ \\ \hline
\end{tabular}
\caption{Optimal measurement configurations for Kuka IIWA R14 820}
\label{configs}
\end{table}

\begin{figure}[H]
\centering
\begin{minipage}[c]{0.5\textwidth}
\centering
    \includegraphics[width=0.3\linewidth]{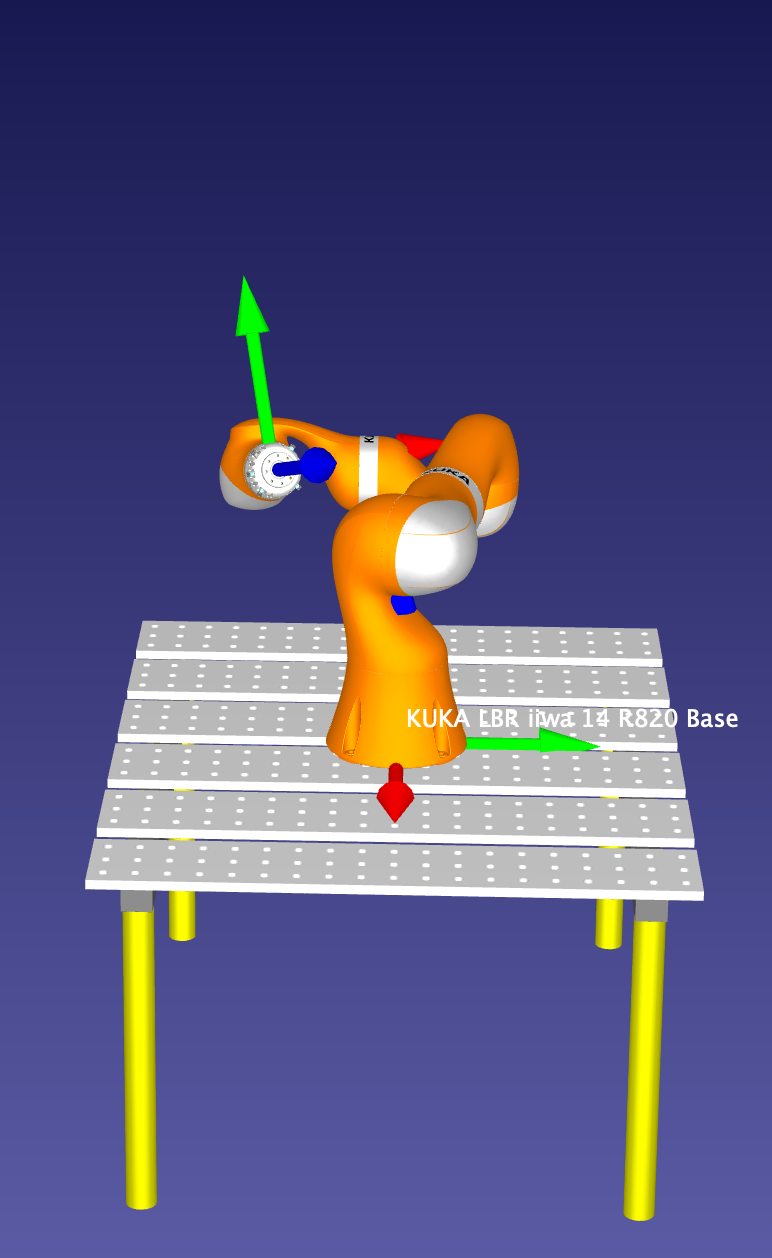}
    \includegraphics[width=0.3\linewidth]{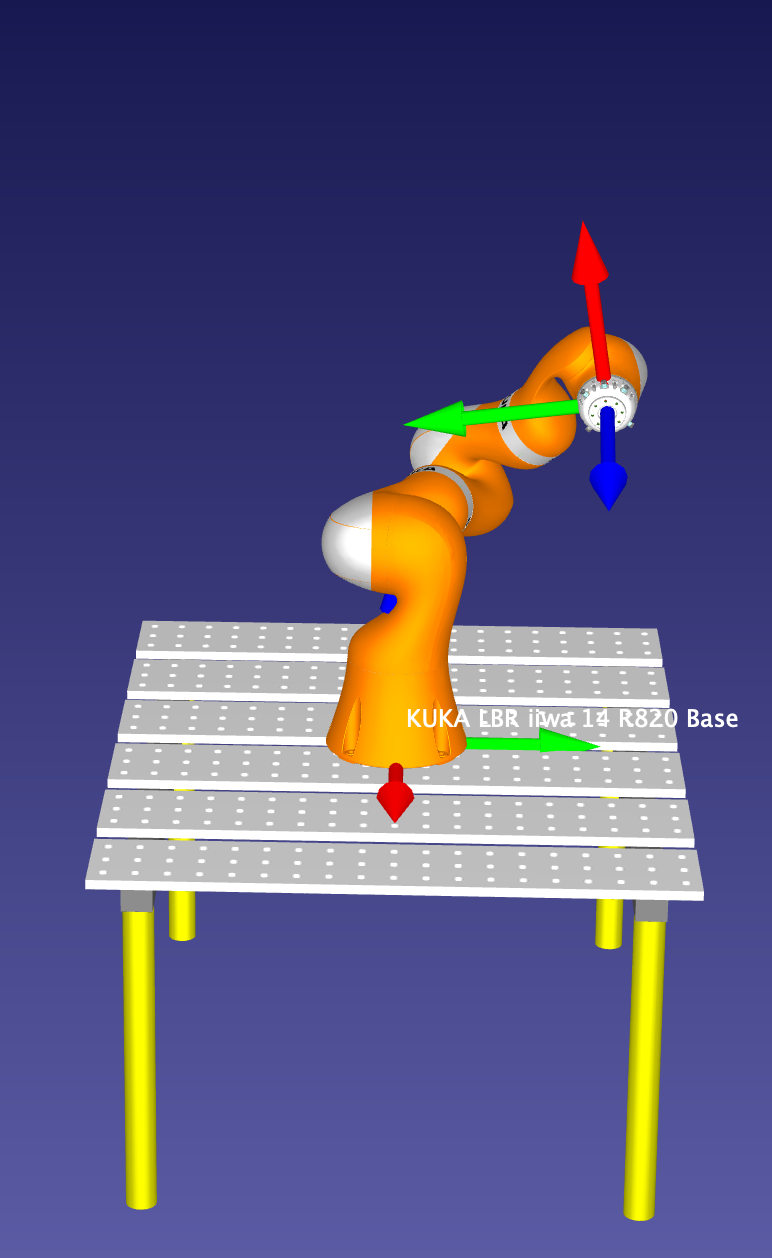}
    \includegraphics[width=0.3\linewidth]{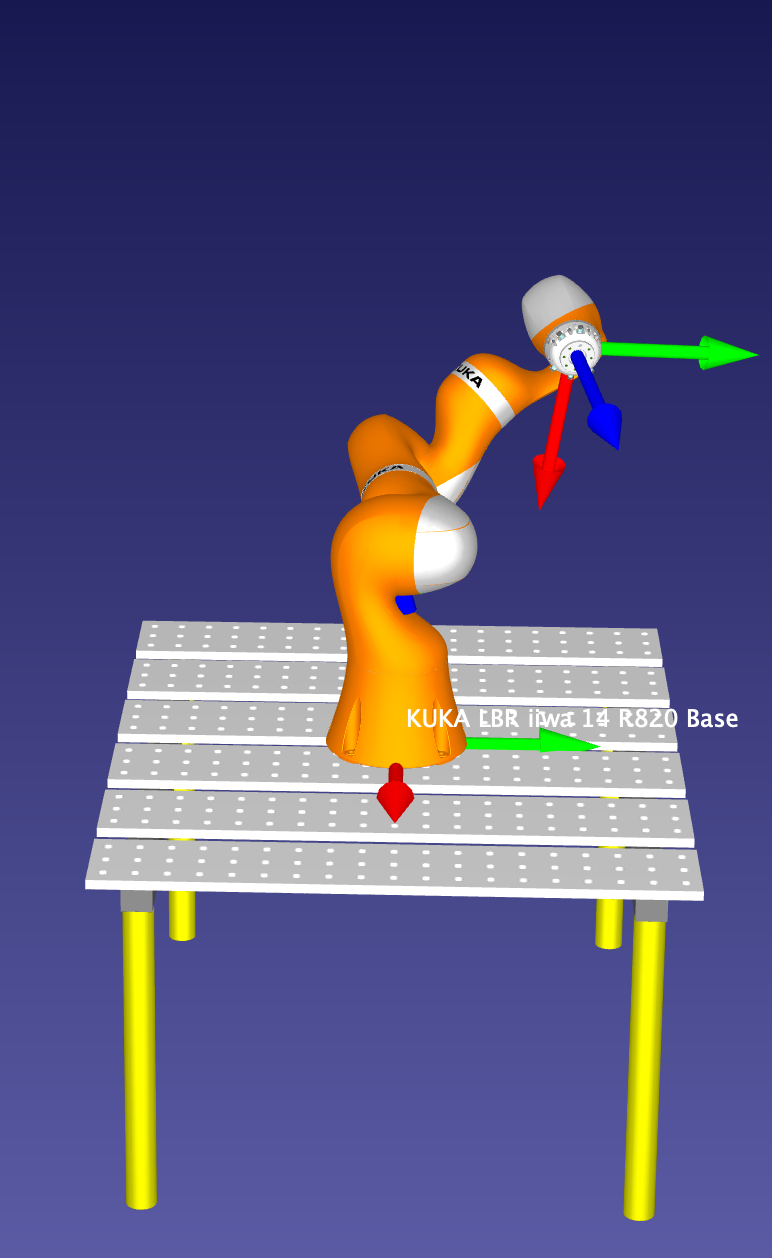}
\end{minipage}
\caption{Three different optimal configurations in the simulator RoboDK}
\label{fig:optimal_conf}
\end{figure}

We simulated the configuration poses to make sure they are safe for the robot and any collision with the surroundings is avoided, Figure \ref{fig:optimal_conf} shows three of these configurations 

We performed the calibration simulation once again using the optimal pose selection, the following figure shows the results obtained (trajectory obtained with optimal and random pose ), We can clearly see that the the optimal plan trajectory follows better the desired trajectory compared to the random plan.

\begin{figure}[H]
    \centering
    \includegraphics[width=0.8\linewidth]{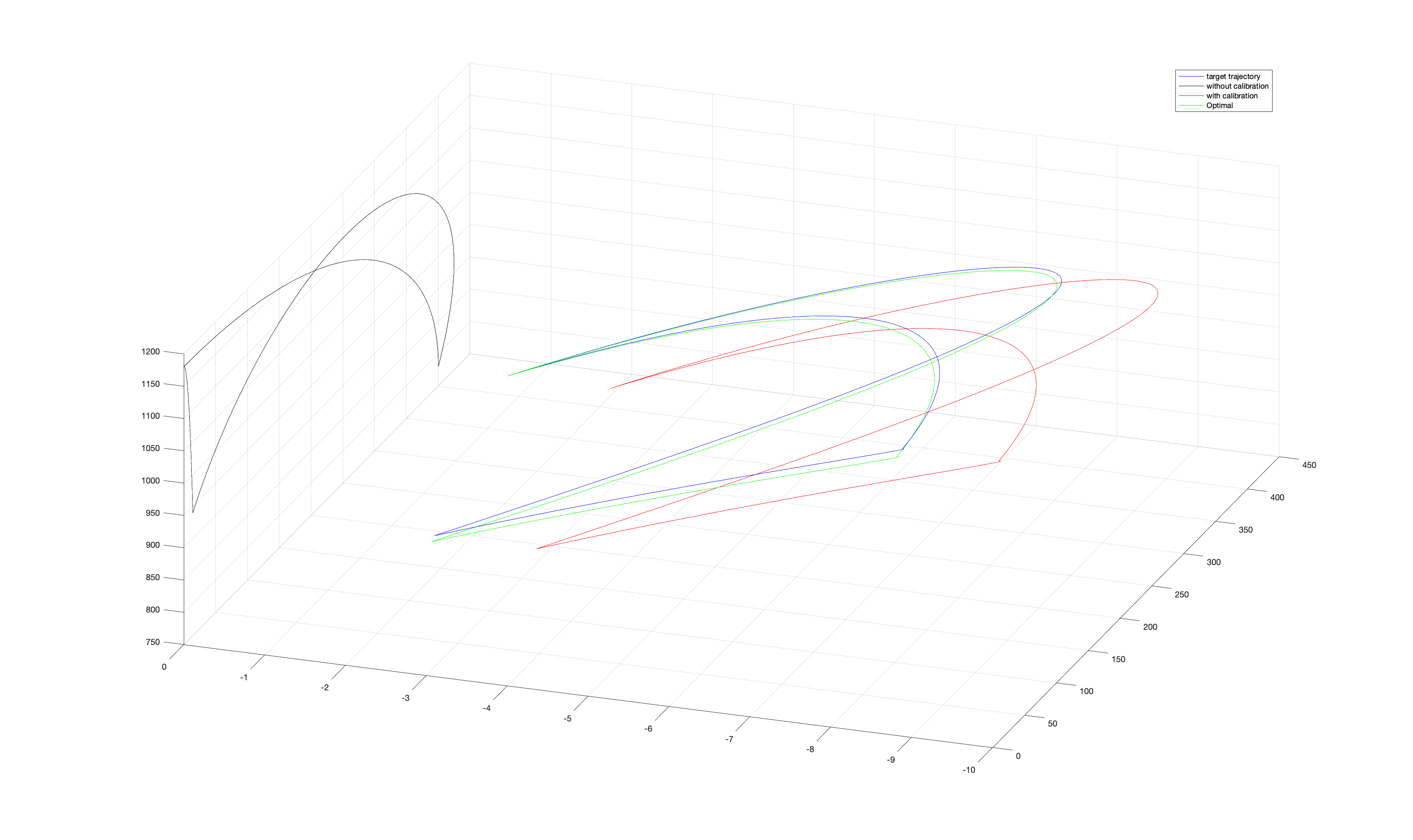}
    \caption{Trajectory obtained before and after calibration (using optimal and random plan) (target trajectory(blue), trajectory without calibration (black), trajectory with random plan calibration (red), trajectory with optimal plan calibration (green)}
    \label{fig:optimal_result}
\end{figure}

Table \ref{table3} shows the identification accuracy for different plans of calibration experiments, we can see that the accuracy with optimal plan is better compared to the random plan, Hence the simulation results confirm the advantages of the calibration using optimal pose selection
\begin{table}[H] 
    \centering
    \begin{tabular}{>{\centering}p{2cm} >{\centering}p{2cm} >{\centering}p{2cm} >{\centering}p{2cm} >{\centering}p{2cm}}
        \multicolumn{5}{c}{\textbf{Design of experiments}} \\
        \hline
        \textbf{Parameter}&  \textbf{Real value} & \textbf{Optimal Plan} &\textbf{Random Plan} & \textbf{Improvement Factor}\tabularnewline
         \hline
         $p_{x1}$&0.0063&0.0003&0.0034&	0.48\tabularnewline
         $p_{y1}$ &0.0081 &  0.0003&  0.0156&0.96\tabularnewline
         $\phi_{y1}$ &-0.0075&2.1908e-06&-0.0074&0.01\tabularnewline
         $\Delta q_2$ & 0.0083&0.00060&0.0083&0.00\tabularnewline
         $p_{y2}$&    0.0026&	-0.0488&	    0.0128&	0.20\tabularnewline
         $p_{2z}$&   415.9754&	415.9700&  415.9723&	0.57\tabularnewline
         $\phi_{y2}$&  -0.0044&	-0.0044	&   -0.0045	&2.41 \tabularnewline
         $\Delta q_3$&     0.0009&	0.0010&   0.0009&0.00\tabularnewline
         $p_{x3}$&     0.0092&	0.0141	 &  -0.0045&	2.82\tabularnewline
         $p_{y3}$&    0.0093&	0.0069&	    0.0088&	0.21\tabularnewline
         $\phi_{y3}$&    -0.0068&	-0.0070	&   -0.0068&	0.00\tabularnewline
         $\Delta q_4$&     0.0094&	0.0094	&    0.0094&	0.00\tabularnewline
         $p_{y4}$&     0.0091&	0.0096	 &   0.0160&	12.90\tabularnewline
         $p_{z4}$&   399.8538&	399.8500&	  399.8661&	3.24\tabularnewline
         $\phi_{y4}$&     0.0060&	0.0062	 &   0.0059&	0.48\tabularnewline
         $\Delta q_5$&    -0.0072&	-0.0072&	   -0.0072&	0.00\tabularnewline
         $p_{x5}$&   -0.0016&	-0.0002	 &  -0.0120&	7.21\tabularnewline
         $p_{y5}$&     0.0083&	0.0146	&    0.0112&	0.46\tabularnewline
         $\phi_{y5}$&     0.0058&	0.0058	&    0.0059&	2.87\tabularnewline
         $\Delta q_6$&     0.0092&	0.0092&	    0.0091&	31.25\tabularnewline
         $p_{y6}$&    0.0031&	0.0093	 &   0.0054&	0.37\tabularnewline
         $p_{z6}$&    -0.0093&	0.0038	&   -1.7009&	129.28\tabularnewline
         $\phi_{y6}$&    0.0070&	0.0069&	    0.0069&	0.67\tabularnewline
         base x & 0.0000	&0.0004&	0.0223&	60.06\tabularnewline
         base y& 0.0000&	0.0003	&    0.0011&	3.56 \tabularnewline
         base z& 364.3399&	0.0000	&  364.3303	&0.00\tabularnewline
         tool 1x&0.0000	&0.0006	& -0.0009&	1.48\tabularnewline
         tool 1y&0.0000&	-0.0488	 &   0.0138	&0.28\tabularnewline
         tool 1z&89.6944&	415.9700&	   91.3873&	0.01\tabularnewline
         tool 2x&0.0000&	-0.0044&	-0.0005&0.11\tabularnewline
         tool 2y&-77.5044 &	0.0010&	  -77.5138&	0.00\tabularnewline
         tool 2z&-44.7472&	0.0141	 & -43.0611&	0.04\tabularnewline
         tool 3x&0.0000&	0.0069	&    0.0084&	1.21\tabularnewline
         tool 3y&78.6892&	-0.0070	  & 78.7003	&0.00\tabularnewline
         tool 3z&-45.4312&	0.0094&	  -43.7463&	0.04\tabularnewline
         \hline
         \multicolumn{4}{c}{Average: } & 7.52\tabularnewline
    \end{tabular}
    \caption{Results of the parameters identification using optimal and random measurement configuration plan}
    \label{table3}
\end{table}

\section{Conclusion:}
The project present the elastostatic modeling and the design of calibration experiment for the spatial anthropomorphic manipulator KUKA iiwa 14 R820, first, for the stiffness modeling we used two approaches to build the cartesian stiffness matrix namely VJM and MSA modeling, then for the calibration, using wise decomposition of the manipulator structure into two planar serial sub-chains, and the properties depicted in \cite{klimchik2014geometrical}, we were able to build 16 measurements configurations describing the optimal pose\\
The complete and irreducible model of the robot was established to perform the calibration simulation, the latter showed clear improvement in the parameters identification with the optimal plan obtained which confirms the efficiency of the approach used.

\bibliographystyle{ieeetr}
\bibliography{bibliography}
\end{document}